\documentclass[10pt,conference]{IEEEtran}
\IEEEoverridecommandlockouts
\usepackage{times}
\usepackage{mathptmx}
\usepackage{courier}
\usepackage{cite}
\usepackage{amsmath,amssymb,amsfonts}
\usepackage{graphicx}
\usepackage{textcomp}
\usepackage{xcolor}
\usepackage{booktabs}
\usepackage{multirow}
\usepackage{array}
\usepackage{url}
\usepackage{balance}
\usepackage{ifthen}
\newboolean{showtodos}
\setboolean{showtodos}{true}
\newcommand{\TODO}[1]{\ifthenelse{\boolean{showtodos}}{\textcolor{red}{\textbf{[TODO: #1]}}}{}}
\title{Edge-Assisted Crowdsensing for Explainable Cyclist Safety Perception}
\author{
\IEEEauthorblockN{Hongsen Cao}
\IEEEauthorblockA{\textit{School of EECS}\\
Queen Mary University of London\\
London, UK\\
hong.cao@qmul.ac.uk}
\and
\IEEEauthorblockN{Chia-Yen Chiang}
\IEEEauthorblockA{\textit{School of EECS}\\
Queen Mary University of London\\
London, UK}
\and
\IEEEauthorblockN{Mona Jaber}
\IEEEauthorblockA{\textit{School of EECS}\\
Queen Mary University of London\\
London, UK}
\and
\IEEEauthorblockN{Meihui Wang}
\IEEEauthorblockA{\textit{School of Public Health}\\
Imperial College London\\
London, UK}
\and
\IEEEauthorblockN{Yasmin Fathy}
\IEEEauthorblockA{\textit{Department of Engineering}\\
University of Cambridge\\
Cambridge, UK}
}
\begin{document}
\maketitle
\begin{abstract}
Cyclist fatality rates remain high despite policy support for active travel, and autonomous vehicles add complexity to mixed traffic. Neither crash statistics nor multi-model detection pipelines scale to continuous monitoring of perceived safety along a route. We present an edge-assisted crowdsensing system: handlebar cameras supply street-level frames, a vision--language model (VLM) turns each frame into structured traffic attributes through a single prompt, and an explainable boosting machine (EBM) combines them with OpenStreetMap road infrastructure and on-bike kinematics to predict a crowdsourced 1--9 safety score. Extraction belongs at the network edge: the fastest self-hosted model needs 2.49\,s per frame and scoring takes 0.4\,ms on a CPU. On 1{,}362 annotated frames from two London routes, the EBM regressor reaches 72.2\% binary accuracy and a mean absolute error of 1.011 under five-fold cross-validation. It is within one accuracy point of black-box ensembles. Its per-feature shape functions give exact attribution: physical lane separation is worth +0.97 rating points, a painted lane +0.70 and a construction site \textminus0.65. Planners can therefore price an infrastructure change in rating points, and the results enter cost-benefit analysis directly. We also report inter-annotator agreement ($\pmb{\alpha}$ = 0.4695), an attribute audit, and a sensitivity analysis linking attribute error to accuracy. With re-encoding, a 20\,s sampling interval serves 1{,}000 riders at 54\,Mbit/s at a cost of 1.4 accuracy points, making continuous monitoring practical.
\end{abstract}
\begin{IEEEkeywords}
cyclist safety, perceived safety, mobile crowdsensing, vision--language models, explainable boosting machine, edge computing, mobile edge generation, spatial cross-validation
\end{IEEEkeywords}
\section{Introduction}
Cycling brings well-documented environmental, health and economic co-benefits~\cite{ding2024}. Yet urban road networks were designed around motorised transport, exposing cyclists to high vehicle volumes and large speed differentials with little physical protection. Autonomous vehicles add further complexity, raising new safety challenges for vulnerable road users~\cite{reyes2022}.

Crash statistics capture only part of the problem. Perceived safety is how safe a cyclist \emph{feels}. It shapes route choice and ridership at least as strongly as objective risk~\cite{cafiso2021,chataway2014}. Questionnaire surveys~\cite{gotschi2018} do not scale to continuous route-level monitoring, and vision pipelines~\cite{costa2025} treat safety as an objective property of the scene. Neither approach measures perceived safety along a route.

Continuous measurement is a mobile crowdsensing problem because the riders carry the sensors and the riders make the judgement. The VLM that describes each scene belongs at the network edge, and downstream stages see only its compact structured output. Mobile edge generation (MEG) formalises this split for generative models: the model runs at the edge and only a sketch is transmitted~\cite{zhonga2024,zhongb2024}. The cost of such a measurement lies in uplink, edge compute and sampling rate, and any saving on them has to be weighed against accuracy. This paper contributes:

\begin{itemize}
  \item \textbf{An edge-deployable crowdsensing pipeline} that turns handlebar video into structured traffic attributes through a single vision--language model (VLM) prompt and fuses them with OpenStreetMap (OSM) road infrastructure and rider kinematics. The prompt replaces a specialised multi-model detection pipeline, and the extraction stage fits in 9.2\,GB of GPU memory on open-weight models, so it can be hosted on a roadside or edge server with only the attribute vector sent onward.
  \item \textbf{An interpretable scoring stage} built on an explainable boosting machine (EBM). It runs in 0.4\,ms on a CPU and can run at any tier of the system. It matches the black-box baselines to within one accuracy point and gives planners per-feature contributions in rating points.
  \item \textbf{A per-stage deployment budget for city-scale operation}, measuring uplink, edge compute and sampling rate: re-encoding cuts the payload 23$\times$, and a 20\,s interval serves a 1{,}000-rider fleet at 54\,Mbit/s at a cost of 1.4 accuracy points.
\end{itemize}
\section{Related Work}
\textbf{Mobile crowdsensing and edge inference.} Mobile crowdsensing treats participant-carried sensing as a systems problem in which coverage, participant burden and uplink cost are design variables~\cite{ganti2011,guo2015}. Karakaya~\textit{et al.}~\cite{karakaya2023} treat cycling-surface sensing this way, splitting processing between rider phones and the cloud. Serving large vision--language models at the network edge is an active research area~\cite{sharshar2025}, and MEG places the generative model itself at the edge so that only compact sketches traverse the network~\cite{zhonga2024,zhongb2024}.

\textbf{Cycling safety assessment.} Recent reviews address cyclist safety broadly~\cite{scarano2023} and perceived safety in particular~\cite{ferreira2022}. Chataway~\textit{et al.}~\cite{chataway2014} find that infrastructure quality shapes how cyclists interpret risk. Questionnaires measure perceived risk under varying conditions~\cite{cafiso2021,gotschi2018}, but they scale poorly and cannot resolve variation along a route. Crash databases~\cite{cyclestreets2024} record outcomes only and cannot separate objective risk from subjective experience.

\textbf{Sensing and vision for scene-level safety.} Street-level imagery has been used to score bikeability~\cite{ito2021}, to rate cycling safety from geospatial and visual cues~\cite{debock2023} and to relate perceived safety to accident records~\cite{ye2024}. Costa~\textit{et al.}~\cite{costa2025} rank environments by pairwise image comparison, and the ranking does not identify which scene elements drive it. Christie~\textit{et al.}~\cite{christie2025} link helmet-mounted video to near-miss frequency, and Chiang~\textit{et al.}~\cite{chiang2024} release an ego-view cyclist dataset; both motivate the on-bike design. Wang~\textit{et al.}~\cite{wang2024} enrich Global Positioning System (GPS) trajectories with road-network context, and we build on the same route-level fusion.

\textbf{Explainable AI in transportation.} Post-hoc saliency methods explain predictions at the pixel level, whereas planners work with per-variable quantities. Generalised additive models (GAMs)~\cite{hastie1986} and their boosted variant the EBM~\cite{lou2013} give inherently interpretable predictions and have been applied to cycling demand~\cite{lee2025}.

\textbf{VLMs for structured scene understanding.} Vision--language pre-training such as Contrastive Language--Image Pre-training (CLIP)~\cite{radford2021} gives strong zero-shot transfer to new visual domains. However, traffic applications have targeted open-ended question answering and narration~\cite{sima2024}, and downstream modelling needs structured extraction. We use structured prompting to extract safety-relevant scene attributes as JavaScript Object Notation (JSON), which replaces a specialised multi-model detection pipeline with a single maintainable query. Section~\ref{sec:features} validates the resulting attributes against human labels.
\section{Methodology}
\begin{figure}[t]
  \centering
  \includegraphics[width=\columnwidth]{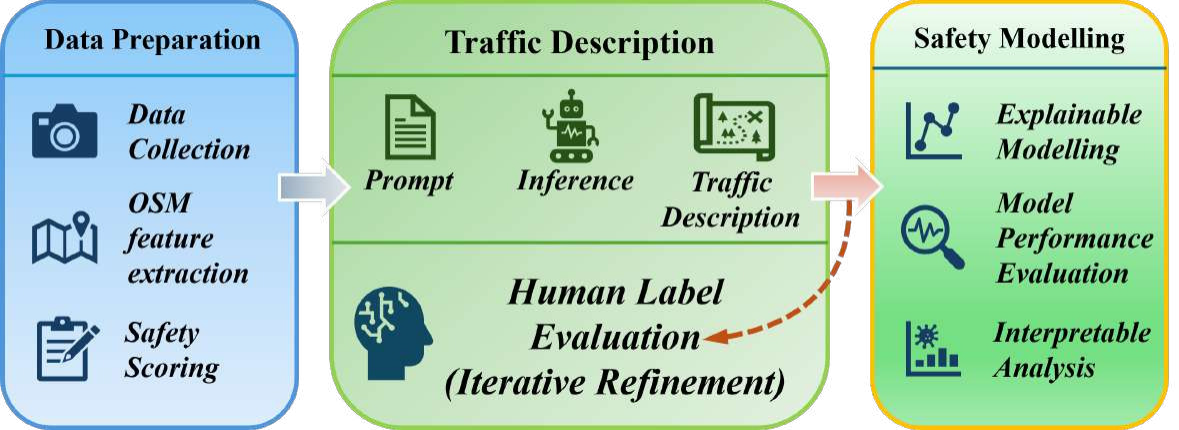}
  \caption{Three-stage pipeline: data preparation, VLM-based traffic description at the network edge, and EBM-based safety modelling with interpretable analysis.}
  \label{fig:workflow}
\end{figure}

Fig.~\ref{fig:workflow} summarises the pipeline in three stages: capture on the bicycle, attribute extraction at the network edge, and scoring wherever the attribute vector is consumed.
\subsection{Data Collection and Safety Scoring}
\label{sec:data}
Two cyclists rode two London routes with handlebar-mounted GoPro cameras that log video, inertial data and GPS fixes. Route~A covers 11\,km near Queen Mary University of London and Route~B 2.9\,km in Farringdon. We sampled one frame every five seconds, yielding 1{,}419 usable frames after quality filtering (1{,}220 from Route~A, 199 from Route~B), each 5312$\times$2988 pixels and a 3.13\,MB payload. Thirty-two recordings span 21.9\,km along-track: 19.1\,km on Route~A over 28 recordings that revisit the same streets, and 2.8\,km on Route~B over four. The 1{,}362 frames carrying a GPS fix form the modelling set, a density of 62.3 frames per rider-kilometre.

We collected annotations on a custom crowdsourcing platform~\cite{cyclingsafety2024}. Ninety-nine annotators rated frames on a 1--9 ordinal scale from \emph{incredibly unsafe} to \emph{incredibly safe}. They were screened for cycling at least three times a week and sampled across age groups. Annotators supplied 7{,}045 ratings over the 1{,}252 frames for which per-annotator records were retained (5.63 ratings per frame on average, median 5, range 1--29); 1{,}165 of those frames were rated twice or more and 87 exactly once. The ground-truth score $y$ for each frame is the mean of its available ratings, and we derive a binary label with a threshold $\tau=5$:
\begin{equation}
  c_i = \begin{cases} 1 & y_i > \tau,\\ 0 & \text{otherwise,}\end{cases}
  \label{eq:label}
\end{equation}
where $c_i=1$ denotes a safe environment. The resulting dataset is near-balanced: of the 1{,}362 frames in the modelling set, 45.2\% are labelled safe and 54.8\% unsafe. Averaged scores have a median of 4.88 and quartiles of 4.00 and 6.14, a slight skew towards the unsafe side. The quartiles bracket $\tau=5$, so the boundary falls inside the middle 50\% of the data and splits the classes more evenly than a cut into either tail.

\textbf{Agreement and rater profile.} Over the multiply-rated frames, Krippendorff's $\alpha$ under an ordinal difference function is 0.4695 with a 95\% confidence interval (CI) of [0.4361, 0.4985], and the median across-annotator variance is 1.25. Agreement is moderate, as expected for a judgement that depends on individual risk tolerance, and it is the reason we model the \emph{mean} rating. Familiarity also shifts ratings. On the 980 frames rated both by annotators who had ridden the location and by annotators who had not, a within-frame paired comparison shows that familiarity \emph{lowers} the rating by 0.121 points (95\% CI [0.031, 0.208]).
\subsection{Multimodal Feature Extraction}
\label{sec:features}
\textbf{VLM-based traffic attributes.} For each frame we issued a single prompt to Grok-2-Vision-1212 requesting a structured JSON response, and the model returned 14 attributes. Five enter the models: cycling lane, lane separation, construction site, immediate crash risk from traffic and high traffic density. The rest cover intersections, traffic signals, buses, pedestrians, surface wetness, parks and visibility, and none of them is a model input. Table~\ref{tab:vlm} compares three VLMs against human-audited ground truth on six visually verifiable attributes. Grok-2-Vision reaches the highest average accuracy (66.7\%) at the lowest per-token cost and supplies every attribute used below.
\begin{table}[t]
  \centering
  \caption{VLM comparison on human-audited attributes (\% accuracy) and per-token price, with the always-absent baseline for the three attributes that carry corpus-wide labels.}
  \label{tab:vlm}
  \resizebox{\columnwidth}{!}{%
  \begin{tabular}{lcccc}
    \toprule
    Attribute & Claude Sonnet & GPT-4o & Grok-2-Vision & Baseline \\
    \midrule
    \texttt{is\_cycling\_lane\_present}        & 68.9 & 55.8 & \textbf{83.6} & 67.0 \\
    \texttt{is\_cycling\_lane\_separation\_present} & 83.8 & 55.0 & \textbf{86.3} & 80.4 \\
    \texttt{is\_intersection}                  & 38.6 & 63.4 & 53.5 & 85.6 \\
    \texttt{is\_construction\_site\_present}   & 59.9 & 57.6 & 58.2 & --- \\
    \texttt{immediate\_crash\_risk\_from\_traffic} & 50.4 & 60.8 & 53.9 & --- \\
    \texttt{is\_high\_traffic\_density}        & 62.3 & 67.1 & 64.9 & --- \\
    \midrule
    Average accuracy                           & 60.7 & 60.0 & \textbf{66.7} & --- \\
    \$/M input tokens                          & 3.00 & 2.50 & 2.00 & --- \\
    \$/M output tokens                         & 15.00 & 10.00 & 10.00 & --- \\
    \bottomrule
    \multicolumn{5}{l}{\footnotesize Baseline: accuracy of always predicting the majority (absent) class,} \\
    \multicolumn{5}{l}{\footnotesize computed on the 1{,}362 frames carrying human labels for that attribute.} \\
  \end{tabular}}
\end{table}

\textbf{Attribute audit.} Three of the six attributes carry corpus-wide human labels on 1{,}362 frames, and comparison against the always-absent baseline in Table~\ref{tab:vlm} shows which of them can be trusted. Cycling lane and lane separation clear their baselines by 16.6 and 5.9 points, and we retain both. Intersection does not: at 53.5\% it falls below its 85.6\% baseline, because the extractor reports an intersection in 57.1\% of frames against a human rate of 14.4\%. We exclude it from every model. The same over-reporting affects the two retained attributes more mildly. The extractor marks them present in 43.2\% and 25.0\% of frames, against human rates of 33.0\% and 19.6\%. We audited the other three on a sample only, so they carry no baseline. Section~\ref{sec:robust} bounds the cost of residual attribute error.

\textbf{OSM road infrastructure.} We map-matched GPS-tagged frames to the OSM road network by projecting each onto the nearest candidate road segment; the largest matched distance was 46.4\,m. We aggregated the OSM road class (\texttt{highway}) to six levels, giving route-level context independent of the frames. The residual \texttt{other} level covers 46.9\% of Route~A.

\textbf{Kinematic features.} GoPro inertial metadata provides three-dimensional speed (\texttt{speed3d}) and acceleration (\texttt{accl3d}), capturing dynamics the frames cannot show. Eight features enter the models: the five attributes named above, the aggregated road class, and the two kinematic channels.
\subsection{Explainable Boosting Machine}
We adopt EBMs~\cite{lou2013}, which extend GAMs~\cite{hastie1986} with gradient boosting over shallow decision trees and a small set of selected pairwise interaction terms:
\begin{equation}
  g\!\left(\mathbb{E}[Y \mid \mathbf{x}]\right) = \beta_0 + \sum_{j=1}^{p} f_j(x_j) + \sum_{i<j} f_{ij}(x_i,x_j).
  \label{eq:ebm}
\end{equation}
Here $Y$ is the target, the score $y\in[1,9]$ for regression and the binary label $c$ for classification. The vector $\mathbf{x}$ holds the $p=8$ features, $\beta_0$ is the intercept, and $g(\cdot)$ is the link, the identity for regression and the logit for classification. Each $f_j$ is a univariate shape function learned by boosting single-feature residuals. The terms $f_{ij}$ capture pairwise effects, with candidate pairs selected by interaction strength~\cite{lou2013}. The regressor minimises mean squared error and the classifier minimises cross-entropy. Each term in Eq.~\eqref{eq:ebm} plots as a shape function that shows how a feature raises or lowers the predicted score, and the attribution is read directly from the fitted terms.

\textbf{Protocol and settings.} We evaluated all models by five-fold cross-validation over the GPS-tagged frames, using \texttt{KFold} with shuffling and \texttt{random\_state}=42. Each fold holds 272--273 test frames. Folds are not stratified, so class balance varies across folds: the safe share spans 0.404--0.484. Seed 1 governs model-internal randomness. We fixed every setting in advance, so no model was tuned against the evaluation data. The configurations are EBM at the defaults of the \texttt{interpret} library; logistic regression \texttt{max\_iter}=1000; decision tree \texttt{max\_depth}=5; random forest 200 trees, \texttt{max\_depth}=8; XGBoost 200 trees, \texttt{max\_depth}=4, learning rate 0.05. The EBM consumes raw features; the remaining models use label encoding with median imputation, and the encoder and imputer are fitted on training folds only.
\section{Results}
\subsection{Exploratory Feature Analysis}
\textbf{Kinematic features.} Perceived safety increases with cycling speed up to approximately 4\,m/s and then declines at higher values. Annotators rated still images and never saw rider speed, so the curve reflects the infrastructure that permits higher speeds.
\begin{figure}[t]
  \centering
  \includegraphics[width=\columnwidth]{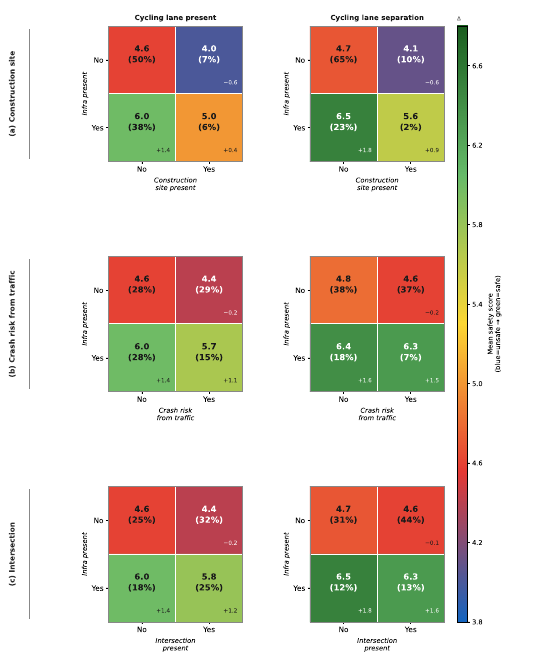}
  \caption{Mean safety score for each combination of cycling infrastructure (column groups: cycling lane present, cycling lane separation) and adverse condition (row panels: construction site, crash risk from traffic, intersection). Each cell gives the mean score and the proportion of observations in brackets. The no-infrastructure, no-hazard cell of each column group is the baseline, and every other cell adds its signed change from it.}
  \label{fig:combos}
\end{figure}

\textbf{Feature combination analysis.} Fig.~\ref{fig:combos} shows mean scores for cycling infrastructure combined with three adverse conditions. Construction sites show the strongest negative contrast, about $-1.0$ points even where physical separation is present. Separation reduces the penalty for immediate crash risk from traffic to $-0.1$ points, compared with $-0.3$ under lane marking alone. A physical barrier reduces perceived danger beyond what lane marking alone achieves. Intersections cost about 0.2 points regardless of infrastructure.

\textbf{Categorical feature ranking.} Ranking the binary attributes by the raw contrast $\Delta=\mathrm{mean}_{\text{Yes}}-\mathrm{mean}_{\text{No}}$ agrees in direction with the combinations above. Cycling lane separation ($\Delta>1.0$) and lane presence show the largest positive contrasts, while construction sites and immediate crash risk show the largest negative ones ($\Delta<-0.4$). These contrasts are marginal; the additive model of Section~\ref{sec:attrib} conditions on the remaining features.
\subsection{Model Performance}
\textbf{Classification.} We compare the EBM with two interpretable baselines, logistic regression and a decision tree, and with two black-box references, a random forest and XGBoost. Table~\ref{tab:models} reports the binary task under random five-fold cross-validation. Among interpretable models the EBM regressor attains the highest accuracy (72.2\%) and weighted F1 (0.719), 5.7 points above logistic regression. It is within 1.0 point of both XGBoost (71.5\%) and random forest (73.2\%). Section~\ref{sec:robust} shows that no model pair separates under a tighter protocol. The EBM regressor's unsafe-class recall (0.815) exceeds its safe-class recall (0.623), so it errs on the side of caution.
\begin{table}[t]
  \centering
  \caption{Model comparison for binary safety classification under random five-fold cross-validation. Differences of about one point at this sample size do not establish an ordering; see Section~\ref{sec:robust}.}
  \label{tab:models}
  \resizebox{\columnwidth}{!}{%
  \begin{tabular}{lccccc}
    \toprule
    Model & Acc & F1(W) & Rec(U) & Rec(S) & MAE \\
    \midrule
    \multicolumn{6}{l}{\textit{Interpretable}} \\
    EBM regressor (ours)  & 0.722 & 0.719 & 0.815 & 0.623 & 1.011 \\
    EBM classifier (ours) & 0.711 & 0.710 & 0.799 & 0.633 & --- \\
    Logistic regression   & 0.665 & 0.664 & 0.761 & 0.580 & --- \\
    Decision tree         & 0.669 & 0.662 & 0.836 & 0.520 & 0.999 \\
    \midrule
    \multicolumn{6}{l}{\textit{Black-box reference}} \\
    Random forest         & 0.732 & 0.732 & 0.813 & 0.660 & 0.988 \\
    XGBoost               & 0.715 & 0.715 & 0.754 & 0.680 & 1.016 \\
    \bottomrule
    \multicolumn{6}{l}{\footnotesize Acc: accuracy. F1(W): weighted F1. Rec(U)/Rec(S): recall for the} \\
    \multicolumn{6}{l}{\footnotesize unsafe/safe class. Mean absolute error} \\
    \multicolumn{6}{l}{\footnotesize (MAE) on the 1--9 scale, regression-capable models only.} \\
  \end{tabular}}
\end{table}

\textbf{Regression.} The EBM regressor achieves a mean absolute error (MAE) of 1.011 and a root-mean-square error of 1.303 on the 1--9 scale, so the average prediction error is about one point. Within-tolerance accuracy reaches approximately 60\% at $\pm1$ and 87\% at $\pm2$. The median annotation variance is 1.25, so two individual raters routinely differ by more than a point. The target is a mean of several ratings and correspondingly less noisy, so $\pm2$ is a practical tolerance.
\subsection{Robustness: Spatial Blocking and Attribute Noise}
\label{sec:robust}
Samples from a moving sensor are spatially autocorrelated. Frames five seconds apart along a route are near-duplicates, so random folds can place almost identical frames in training and test. To quantify how much this inflates accuracy, we re-ran every model under a second five-fold protocol differing only in grouping. We cut each recording into contiguous 200\,m along-track blocks, 124 in total, and formed folds with \texttt{GroupKFold} so that no block spans the train/test boundary; we call this spatial blocking. We assigned blocks from coordinates alone, before fitting any transformation. Neither protocol stratifies, so grouping is the only manipulated difference. Spatial blocking widens the per-fold safe share to 0.335--0.531, against 0.404--0.484 under random folds.

Measured accuracy is higher under random folds for five of the six models. Under frame-paired comparison, the gain is 3.7 points for random forest (95\% CI [1.8, 5.7], Holm $p<0.001$) and 2.6 points for the EBM regressor ([0.8, 4.7], $p=0.004$). The remaining gains are too small for the test to resolve, and logistic regression shows none, since a linear model cannot exploit near-duplicate neighbours. Under spatial blocking, no pair of models separates in any of the 15 exact McNemar tests with Holm correction. The ordering in Table~\ref{tab:models} is therefore not a ranking, and its accuracy level holds for the random-fold protocol that produced it.

\textbf{An end-to-end baseline.} On the same spatially blocked folds, a linear probe on CLIP ViT-L/14 embeddings does not separate from the EBM (Holm $p=0.3676$ over seven regularisation settings), so the comparison is unresolved at this sample size. Its best setting leads by 2.6 points, and that setting was chosen on the evaluation folds. The probe also has no additive attributions.

\textbf{Attribute noise.} Corrupting the five VLM inputs at rate $p$ and retraining under spatial blocking degrades accuracy roughly linearly, by about 3.3 points per ten points of added attribute error, reaching 9.8 points at $p=0.30$. The five inputs average 69.4\% accuracy in Table~\ref{tab:vlm}, so their 30.6\% error rate is already at the top of the swept range. Attribute quality is the main limit on accuracy here. The model tolerates that error because it combines several weak, partly independent cues and depends on none of them alone.
\subsection{Feature Importance and Explainability}
\label{sec:attrib}
\begin{figure}[t]
  \centering
  \includegraphics[width=\columnwidth]{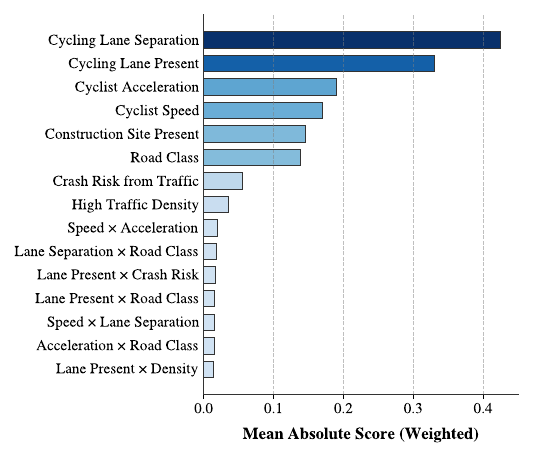}
  \caption{Global feature importance from the EBM, as the mean absolute weighted score. Every main-effect term outranks every pairwise interaction term.}
  \label{fig:importance}
\end{figure}

As Fig.~\ref{fig:importance} shows, individual shape functions dominate pairwise interaction terms in mean absolute contribution. In descending order, the five most influential features are physically separated cycling lane, painted cycling lane, cyclist acceleration, cyclist speed and construction site.

On the rating scale, physical separation is worth $+0.97$ points and a painted lane $+0.70$, while a construction site costs 0.65 points. A construction site reduces the benefit of protective infrastructure but does not cancel it. Road class has the widest marginal contrast of any categorical feature ($\Delta=1.52$), but ranks below these inside the model. Across 150 bootstrap resamples of the 124 spatial blocks, the two cycling-lane terms appear in the top three every time. Road class, construction and speed exchange positions, so we treat those three as one tier.
\section{Deployment Budget: Uplink, Compute and Sampling}
\label{sec:deploy}
City-scale monitoring is constrained less by accuracy than by uplink and compute cost. Scoring a frame with the trained EBM takes 0.4\,ms on CPU; the fastest VLM configuration needs 2.49\,s on an A100-80GB, a factor of roughly 6{,}000. Extraction should therefore be offloaded, as in mobile edge computing~\cite{mao2017} and in recent work on vision--language inference at the edge~\cite{sharshar2025}. In MEG terms the attribute vector is the sketch, and downstream stages work from it without running the heavy model~\cite{zhonga2024}. In the proposed deployment each frame enters the edge as a 0.13\,MB re-encoded image and leaves it as a 14-field JSON record.
\begin{table}[t]
  \centering
  \caption{Self-hosted extraction on a 309-frame stratified subset (A100-80GB, bf16, batch 1, 1024\,px long edge). Accuracies are on cycling lane and lane separation, the two attributes the audit retains.}
  \label{tab:selfhost}
  \resizebox{\columnwidth}{!}{%
  \begin{tabular}{lcccc}
    \toprule
    Model & Lat.\ (s) & VRAM & Frames/GPU\,h & Lane / Sep. \\
    \midrule
    Qwen3-VL 4B   & 2.74 & 9.2\,GB  & 1312 & 0.854 / 0.816 \\
    Qwen3-VL 8B   & 2.99 & 17.8\,GB & 1205 & 0.819 / 0.816 \\
    Qwen3-VL 32B  & 8.94 & 67.2\,GB & 403  & 0.767 / 0.822 \\
    InternVL3 8B  & 2.49 & 16.1\,GB & 1444 & 0.877 / 0.825 \\
    Llama-3.2 11B & 3.63 & 22.3\,GB & 992  & 0.796 / 0.793 \\
    \bottomrule
    \multicolumn{5}{l}{\footnotesize Lat.: mean latency per frame. VRAM: peak GPU memory.} \\
    \multicolumn{5}{l}{\footnotesize Frames/GPU\,h: sustained throughput. Lane / Sep.: accuracy on} \\
    \multicolumn{5}{l}{\footnotesize cycling lane and lane separation.} \\
  \end{tabular}}
\end{table}

\textbf{Self-hosted extraction.} A commercial endpoint charges per frame and requires rider video and GPS to leave the operator's control. Table~\ref{tab:selfhost} benchmarks five open-weight VLMs. Accuracy on cycling lane is comparable to the commercial extractor's, and accuracy does not rise with model size. The 4B model exceeds the 32B by 8.7 points on cycling lane (Holm $p=0.0036$) at 3.3$\times$ the speed in one-seventh the memory. On lane separation all five are close to the 80.4\% always-absent baseline of Table~\ref{tab:vlm}, so only the cycling-lane column discriminates. Fifteen paired tests cover five models and the three attributes with corpus-wide labels at both input resolutions. The 1024\,px input is favoured in seven and native capture in none. These extractors suit a new deployment; checking equivalent downstream attribution would require re-extracting the corpus.

\begin{figure}[t]
  \centering
  \includegraphics[width=0.92\columnwidth]{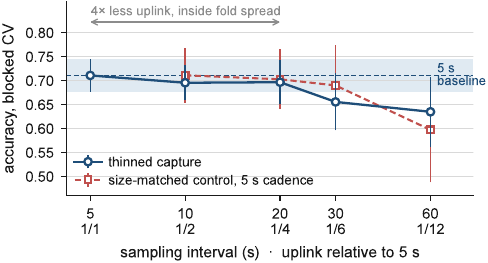}
  \caption{Sampling interval against accuracy under spatial blocking, with the uplink each interval implies relative to 5\,s capture on the same axis. Shading marks the fold-to-fold spread of the 5\,s baseline. The size-matched control keeps the 5\,s cadence while using the same number of frames as the thinned setting. Bars span folds.}
  \label{fig:sampling}
\end{figure}

\textbf{Uplink and sampling interval.} At 3.13\,MB per frame and one frame every 5\,s, raw capture implies 5.0\,Mbit/s per rider, so a 1{,}000-rider fleet would demand 5.0\,Gbit/s. Re-encoding to a 1024\,px long edge at JPEG quality 85 cuts the payload by 23$\times$, and a 20\,s interval removes another factor of four, bringing the fleet requirement to 54\,Mbit/s. Fig.~\ref{fig:sampling} sweeps the interval from 5\,s to 60\,s against the uplink it implies. Accuracy under spatial blocking stays inside the fold-to-fold spread of the 5\,s baseline out to 20\,s, where the uplink is already four times smaller, and declines from 30\,s. The 20\,s setting costs 1.4 accuracy points. A size-matched control at the 5\,s cadence tracks the thinned curve over the same range. Compute then becomes the limit: the same fleet emits 180{,}000 frames per hour, and extracting them at the fastest measured throughput would occupy about 125 A100-class accelerators continuously. At city scale, extraction is a datacentre-class workload at the network edge; the 0.4\,ms scoring step can run wherever the attribute vector arrives, including the bicycle.
\section{Discussion and Limitations}
\textbf{Experience weighting.} Weighting ratings 0.8 to 0.2 in favour of annotators who had ridden the location improves regression MAE from 1.011 to 0.95 and raises accuracy by about two points, but reduces safe-class recall by roughly ten points. It also changes the target scores, so the two are not scored on identical labels; we report the unweighted model as primary.

\textbf{Limitations.} The VLM processes each frame independently and may miss cross-frame context; contrastive prompting across consecutive frames could address this. The dataset covers two routes in inner London, and cross-city generalisation remains to be demonstrated. Annotators cycle at least three times weekly, and familiarity alone moves a rating by 0.121 points, so the model reflects the judgement of frequent local riders. Extending the annotator pool to occasional and non-cyclists is the next step.
\section{Conclusion}
We presented an edge-assisted crowdsensing system that turns handlebar video into a structured sketch of each scene at the network edge, fuses it with OSM road infrastructure and rider kinematics, and scores perceived safety with an explainable boosting machine. The EBM regressor achieves 72.2\% binary accuracy and MAE 1.011 on a 1--9 scale. It stays within one point of the black-box ensembles under random folds, and under spatial blocking no model pair separates. With scoring some 6{,}000$\times$ cheaper than extraction and a 20\,s interval serving a 1{,}000-rider fleet at 54\,Mbit/s, the pipeline is a practical basis for continuous safety monitoring. The system gives planners a scale of perceived safety on which to weigh candidate interventions. Neither the structured prompt nor the additive model is specific to cycling, so the same edge-assisted pattern extends to other street-level qualities judged by the people who use the street.
\balance
\bibliographystyle{IEEEtran}

\begin{thebibliography}{00}
\bibitem{ding2024}
D. Ding \textit{et al.}, ``The co-benefits of active travel interventions
beyond physical activity: a systematic review,''
\textit{Lancet Planetary Health}, vol.~8, no.~10, pp.~e790--e803, 2024.
\bibitem{reyes2022}
A. Reyes-Mu\~{n}oz and J. Guerrero-Ib\'{a}\~{n}ez,
``Vulnerable road users and connected autonomous vehicles interaction:
A survey,'' \textit{Sensors}, vol.~22, no.~12, p.~4614, 2022.
\bibitem{cafiso2021}
S. Cafiso, G. Pappalardo, and N. Stamatiadis, ``Observed risk and user
perception of road infrastructure safety assessment for cycling mobility,''
\textit{Infrastructures}, vol.~6, no.~11, p.~154, 2021.
\bibitem{chataway2014}
E.~S. Chataway \textit{et al.}, ``Safety perceptions and reported behaviour
related to cycling in mixed traffic: A comparison between Brisbane and
Copenhagen,'' \textit{Transportation Research Part F}, vol.~23, pp.~32--43,
2014.
\bibitem{gotschi2018}
T. G\"{o}tschi \textit{et al.}, ``Towards a comprehensive safety evaluation
of cycling infrastructure including objective and subjective measures,''
\textit{Journal of Transport \& Health}, vol.~8, pp.~44--54, 2018.
\bibitem{costa2025}
M. Costa \textit{et al.}, ``Which cycling environment appears safer? Learning
cycling safety perceptions from pairwise image comparisons,''
\textit{IEEE Transactions on Intelligent Transportation Systems}, vol.~26,
pp.~1689--1700, 2025.
\bibitem{zhonga2024}
R. Zhong, X. Mu, Y. Zhang, M. Jaber, and Y. Liu, ``Mobile edge generation:
A new era to 6G,'' \textit{IEEE Network}, vol.~38, no.~5, pp.~47--55, 2024.
\bibitem{zhongb2024}
R. Zhong, X. Mu, M. Jaber, and Y. Liu, ``A mobile edge generation approach,''
in \textit{Proc. IEEE Global Communications Conf. (GLOBECOM)}, 2024, pp.~4173--4178.
\bibitem{ganti2011}
R.~K. Ganti, F. Ye, and H. Lei, ``Mobile crowdsensing: current state and
future challenges,'' \textit{IEEE Communications Magazine}, vol.~49, no.~11,
pp.~32--39, 2011.
\bibitem{guo2015}
B. Guo \textit{et al.}, ``Mobile crowd sensing and computing: the review of an
emerging human-powered sensing paradigm,'' \textit{ACM Computing Surveys},
vol.~48, no.~1, pp.~1--31, 2015.
\bibitem{karakaya2023}
A.-S. Karakaya, L. Thomas, D. Koljada, and D. Bermbach, ``A crowdsensing
approach for deriving surface quality of cycling infrastructure,'' in
\textit{Proc. 11th IEEE Int. Conf. Cloud Engineering (IC2E)}, 2023.
\bibitem{sharshar2025}
A. Sharshar, L.~U. Khan, W. Ullah, and M. Guizani, ``Vision-language models
for edge networks: A comprehensive survey,'' 2025, arXiv:2502.07855.
\bibitem{scarano2023}
A. Scarano, M. Aria, F. Mauriello, M.~R. Riccardi, and A. Montella,
``Systematic literature review of 10 years of cyclist safety research,''
\textit{Accident Analysis \& Prevention}, vol.~184, p.~106996, 2023.
\bibitem{ferreira2022}
M.~C. Ferreira \textit{et al.}, ``Identifying the determinants and understanding
their effect on the perception of safety, security and comfort by pedestrians
and cyclists,'' \textit{Transportation Research Part F}, vol.~91,
pp.~136--163, 2022.
\bibitem{cyclestreets2024}
CycleStreets, ``BikeData: Cycling Data Map,''
2024. [Online]. Available: \url{https://bikedata.cyclestreets.net/}
\bibitem{ito2021}
K. Ito and F. Biljecki, ``Assessing bikeability with street view imagery and
computer vision,'' \textit{Transportation Research Part C}, vol.~132,
p.~103371, 2021.
\bibitem{debock2023}
J. De Bock and S. Verstockt, ``Road cycling safety scoring based on geospatial
analysis, computer vision and machine learning,''
\textit{Multimedia Tools and Applications}, vol.~82, pp.~8359--8380, 2023.
\bibitem{ye2024}
Y. Ye, C. Zhong, and E. Suel, ``Unpacking the perceived cycling safety of road
environment using street view imagery and cycle accident data,''
\textit{Accident Analysis \& Prevention}, vol.~205, p.~107677, 2024.
\bibitem{christie2025}
N. Christie \textit{et al.}, ``Spatiotemporal and behavioural correlates of
cycling near misses: evidence from helmet-mounted video data,''
\textit{Accident Analysis \& Prevention}, vol.~224, p.~108272, 2025.
\bibitem{chiang2024}
C.-Y. Chiang \textit{et al.}, ``AllTheDocks road safety dataset: A cyclist's
perspective and experience,'' in \textit{Proc. IEEE 99th Vehicular Technology
Conf. (VTC2024-Spring)}, 2024, pp.~1--5.
\bibitem{wang2024}
M. Wang \textit{et al.}, ``CycleTrajectory: an end-to-end pipeline for
enriching and analysing GPS trajectories to understand cycling behaviour,''
in \textit{Proc. 2nd ACM SIGSPATIAL Workshop on Sustainable Urban Mobility},
2024, pp.~27--32.
\bibitem{hastie1986}
T. Hastie and R. Tibshirani, ``Generalized additive models,''
\textit{Statistical Science}, vol.~1, no.~3, pp.~297--318, 1986.
\bibitem{lou2013}
Y. Lou, R. Caruana, J. Gehrke, and G. Hooker, ``Accurate intelligible models
with pairwise interactions,'' in \textit{Proc. 19th ACM SIGKDD Int. Conf.
Knowledge Discovery and Data Mining}, 2013, pp.~623--631.
\bibitem{lee2025}
E.~H. Lee, ``Understanding the gender gap in bike-sharing services via
explainable artificial intelligence,'' \textit{Transportation Research Record},
vol.~2679, pp.~622--633, 2025.
\bibitem{radford2021}
A. Radford \textit{et al.}, ``Learning transferable visual models from
natural language supervision,'' in \textit{Proc. 38th Int. Conf. Machine
Learning (ICML)}, 2021, pp.~8748--8763.
\bibitem{sima2024}
C. Sima \textit{et al.}, ``DriveLM: Driving with graph visual question
answering,'' in \textit{Proc. European Conf. Computer Vision (ECCV)}, 2024.
\bibitem{cyclingsafety2024}
Digital Twins for Sustainable Development Goals, QMUL,
``CyclingSafety.co.uk: Crowdsourced Cyclist Safety Perception Platform,''
2024.
\bibitem{mao2017}
Y. Mao, C. You, J. Zhang, K. Huang, and K.~B. Letaief, ``A survey on mobile
edge computing: the communication perspective,'' \textit{IEEE Communications
Surveys \& Tutorials}, vol.~19, no.~4, pp.~2322--2358, 2017.
\end{thebibliography}

\end{document}